\pdfoutput=1

\documentclass[10pt,journal,compsoc]{IEEEtran}
\usepackage{amsmath,amsfonts}
\usepackage{algorithmic}
\usepackage{algorithm}
\usepackage{array}
\usepackage[caption=false,font=normalsize,labelfont=sf,textfont=sf]{subfig}
\usepackage{textcomp}
\usepackage{stfloats}
\usepackage{url}
\usepackage{verbatim}
\usepackage{graphicx}
\usepackage{cite}
\usepackage{fancyvrb}
\usepackage[dvipsnames]{xcolor}
\usepackage{multicol}
\usepackage{multirow}
\usepackage{tabularx}
\usepackage{ragged2e}
\usepackage{hyperref}

\begin{document}

\title{3D Object Detection from Point Cloud via \\ Voting Step Diffusion}

\author{Haoran Hou, Mingtao Feng$^\ddagger$, Zijie Wu$^\ddagger$, Weisheng Dong, Qing Zhu, Yaonan Wang and Ajmal Mian
\thanks{Haoran Hou, Zijie Wu, Qing Zhu and Yaonan Wang are with the College of Electrical and Information Engineering, Hunan University, Changsha 410082, China (email: haoranh@stu.xidian.edu.cn; wuzijieeee@hnu.edu.cn; zhuqing@hnu.edu.cn; yaonan@hnu.edu.cn).}

\thanks{Mingtao Feng and Weisheng Dong are with the School of Artificial Intelligence, Xidian University, Xi’an 710071, China (email: mintfeng@hnu.edu.cn; wsdong@mail.xidian.edu.cn).} 

\thanks{Ajmal Mian is with the Department of Computer Science and Software Engineering, The University of Western Australia, Perth, Crawley, WA 6009, Australia (e-mail: ajmal.mian@uwa.edu.au).}

\thanks{$^\ddagger$ denotes corresponding authors.}
}

\markboth{Journal of \LaTeX\ Class Files,~2024}%
{Shell \MakeLowercase{\textit{et al.}}: A Sample Article Using IEEEtran.cls for IEEE Journals}

\IEEEpubid{0000--0000/00\$00.00~\copyright~2023 IEEE}

\IEEEtitleabstractindextext{%

\begin{abstract}
\justifying
3D object detection is a fundamental task in scene understanding. Numerous research efforts have been dedicated to better incorporate Hough voting into the 3D object detection pipeline. However, due to the noisy, cluttered, and partial nature of real 3D scans, existing voting-based methods often receive votes from the partial surfaces of individual objects together with severe noises, leading to sub-optimal detection performance. In this work, we focus on the distributional properties of point clouds and formulate the voting process as generating new points in the high-density region of the distribution of object centers. To achieve this, we propose a new method to move random 3D points toward the high-density region of the distribution by estimating the score function of the distribution with a noise conditioned score network. Specifically, we first generate a set of object center proposals to coarsely identify the high-density region of the object center distribution. To estimate the score function, we perturb the generated object center proposals by adding normalized Gaussian noise, and then jointly estimate the score function of all perturbed distributions. Finally, we generate new votes by moving random 3D points to the high-density region of the object center distribution according to the estimated score function. Extensive experiments on two large scale indoor 3D scene datasets, SUN RGB-D and ScanNet V2, demonstrate the superiority of our proposed method. The code will be released at \url{https://github.com/HHrEtvP/DiffVote}.
\end{abstract}
\vspace{-3mm}
}

\maketitle
\IEEEdisplaynontitleabstractindextext
\IEEEpeerreviewmaketitle

\section{Introduction}
\IEEEPARstart{3}{D} object detection aims to localize oriented 3D bounding boxes and predict associated semantic labels of objects from a 3D point set. As a fundamental technique for 3D scene understanding, 3D object detection is widely used in various downstream tasks, such as autonomous driving~\cite{wu2022sparsedrive, chen2022focaldrive}, mobile robots~\cite{ten2017grasp, zhao2021graspregnet}, and high-level semantic SLAM (Simultaneous Localization and Mapping)~\cite{fan2022blitz, chen2019suma++}. 
Compared to image based object detection, 3D object detection provides the exact 3D location, orientation and size of objects.
However, 3D object detection is relatively more challenging due to the noisy, cluttered, and partial nature of 3D point clouds. This also makes it very difficult to apply the high-performance pipelines used for 2D object detection to 3D point clouds~\cite{rukhovich2022fcaf3d}.

\begin{figure}[t]
\centering
\includegraphics[width=\linewidth]{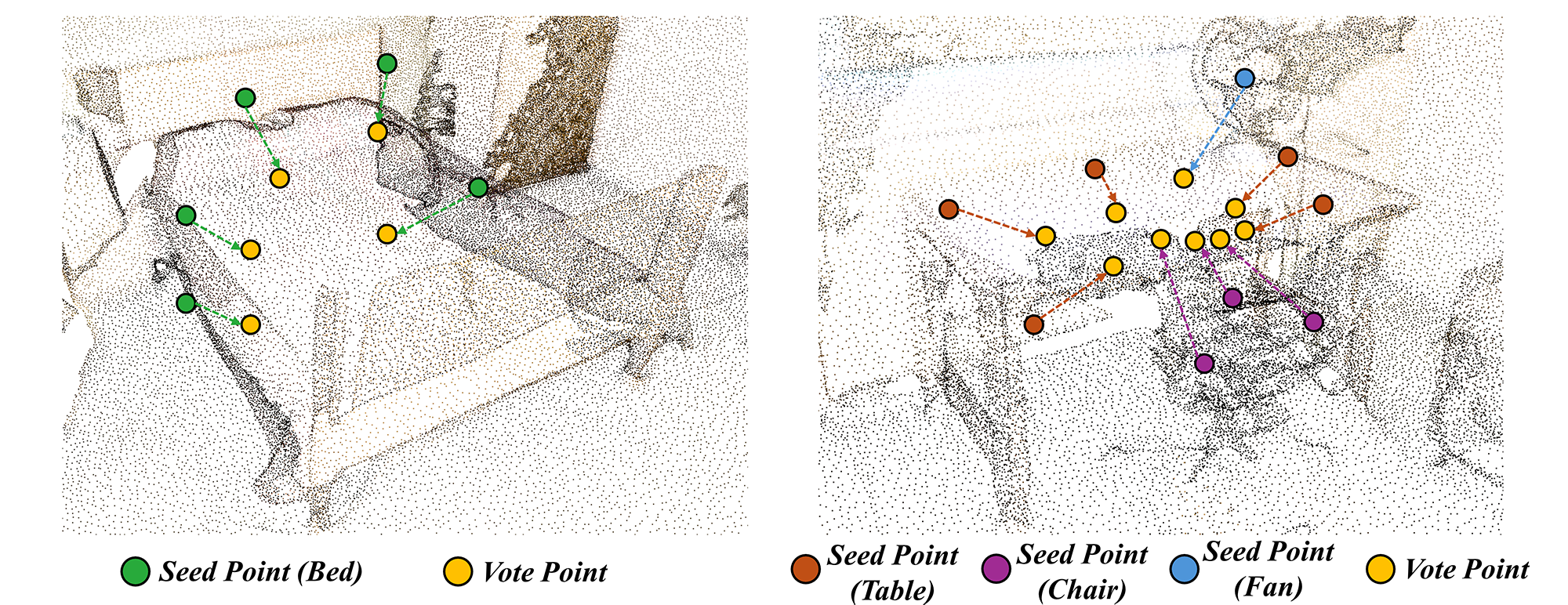}
\vspace{-0.7cm}
\caption{The votes generated by VoteNet usually suffer from partial coverage of the object surfaces and outliers from the cluttered background and adjacent objects.}
\label{fig_problem}
\vspace{-4mm}
\end{figure}

Based on point representation, existing 3D object detection methods can be classified into two categories: grid-based methods and point-based methods. While grid-based methods convert the irregular point clouds to regular data structures such as 3D voxels~\cite{shi2020pv, shi2023pv++}, point-based methods take raw point clouds as input and process point clouds efficiently and robustly~\cite{qi2017pointnet, qi2017pointnet++, TIPrelation}. 
The popular VoteNet~\cite{qi2019votenet} point-wise 3D detector yields outstanding results by integrating Hough Voting into 3D object detection. The voting process is formulated as center point regression and implemented via a multi-layer perception (MLP). These votes are then clustered and aggregated to generate object proposal features, which are used to classify objects and regress their locations.
However, the quality of votes is far from satisfactory due to the complexity and diversity of objects in real 3D scenes. As illustrated in Figure~\ref{fig_problem}, the voting process fails to generate high-quality votes as vote clusters not only fail to uniformly cover the underlying objects but also contain outliers from the background and adjacent objects, making it challenging to detect objects based on these votes~\cite{cheng2021back}.

We argue that generative models can improve the quality of votes and the performance of point-based 3D detectors.
Voting, as the essence in VoteNet, generates new points that lie close to object centers. Because of the distributional properties of point clouds, a point cloud can be modeled as a set of samples from some 3D distribution~\cite{luo2021diffusion}. Therefore, the voting process amounts to generating new samples in the high-density region of the distribution of object centers. 
While MLPs used in VoteNet and voting-based methods~\cite{xie2021venet, xie2020mlcvnet, wang2022rbgnet, cheng2021back} are simpler and faster to train, they ignore the distribution of object centers and are unable to express the relationship between point cloud and object centers when the underlying distribution is complex or uncertain. 
In contrast, generative models are capable of directly modeling the distribution of object centers. This enables the generative models to generate accurate votes in the high-density region of the distribution. More importantly, generative models can learn more domain knowledge and facilitate the application of probability theory, which makes them more robust to the irregular and noisy point cloud.

\begin{figure}[t]
\centering
\includegraphics[width=\linewidth]{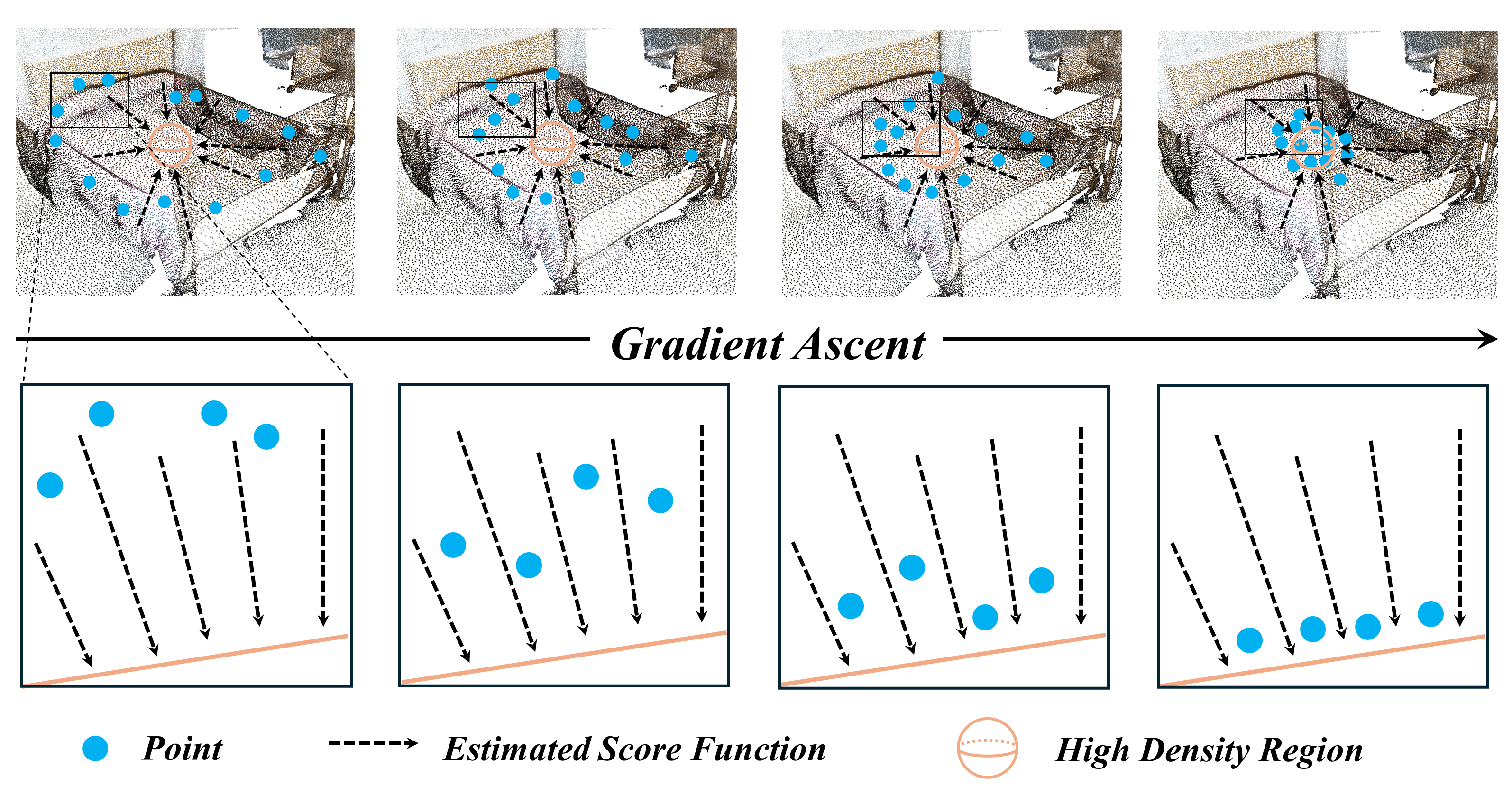}
\vspace{-8mm}
\caption{Illustration of our proposed method. We first estimate the score function of the distribution of object centers. Then, we perform gradient ascent to move points to the high-density region of the distribution. }
\label{fig_overview}
\vspace{-4mm}
\end{figure}

Motivated by the advantages of generative models, we explore on designing a better paradigm of voting and 3D object detection that leverages the power of generative models. Recently, diffusion models~\cite{sohl2015originalddpm, ho2020ddpm} have achieved great success in various vision tasks, such as image synthesis~\cite{rombach2022ldm, vahdat2022lion, nichol2021glide, dhariwal2021diffusionbeatsgan}, semantic segmentation~\cite{amit2021segdiff, baranchuk2021label}, and 3D shape generation~\cite{luo2021diffusion, wu2023sketch, nakayama2023difffacto}. 
In particular, Noise Conditioned Score Networks (NCSNs), a sub-category of diffusion models, generate new samples by gradually moving a random initial sample to the high-density region based on the estimated score function of the target data distribution, where the score function is defined as the gradient of the log density of the data distribution~\cite{song2019generative}.
According to this observation, vote generation can be formulated as moving random 3D points to the high-density region of the distribution of object centers, which can be realized by performing gradient ascent on the score function estimated by noise conditioned score network, as illustrated in Figure~\ref{fig_overview}.
However, to the best of our knowledge, no existing work has adopted noise conditioned score network for 3D object detection. Applying noise conditioned score network to 3D point clouds is not trivial since 3D objects in point clouds are highly cluttered and close to each other, leading to inaccurate score estimation and low recall rates. 

To address the above problems, we present a novel 3D object detection framework that tackles 3D object detection task with noise conditioned score network by directly modeling the distribution of object centers and estimating the score function of the distribution to move random points toward the vicinity of object centers.
Firstly, we generate a set of object center proposals to coarsely identify the high-density region of the distribution. Gaussian noise controlled by a variance schedule is then added to the generated object center proposals to obtain the perturbed object center proposals. Secondly, we utilize a multi-scale score estimation module to estimate the score function of the distribution. Our score estimation module first extracts the fine-grained details and multi-scale information of the perturbed center proposals and then estimates the score function of the distribution by predicting the added Gaussian noise. Finally, based on the score estimates, we move the perturbed object center proposals toward the high-density region of the distribution, aggregate the scene context from the denoised object center proposals, and predict the 3D bounding boxes.

Our main contributions are: 
1) We design a novel paradigm of voting and 3D object detection that leverages the power of noise conditioned score network to generate accurate votes and object proposals. To the best of our knowledge, this is the first work that successfully applies noise conditioned score network to 3D object detection. 
2) To further release the power of noise conditioned score network, we propose a multi-scale score estimation module that efficiently extracts the multi-scale feature of the perturbed data and accurately predicts the added random noise.
3) We design a score-aware object proposal module to remove the added random noise and aggregate the multi-scale features of object center proposals. Results show that our method consistently boosts 3D object detection performance and outperforms existing point-based methods.

\vspace{-2mm}
\section{Related Works}
\subsection{3D Object Detection.}
\noindent Due to the noisy, irregular, and partial nature of 3D point clouds, 3D object detection remains a challenging task. Existing works can be classified into two categories based on the point cloud representation, i.e., grid-based and point-based. Grid-based methods convert point clouds into regular data, such as 2D bird’s eye view images~\cite{li2023bevdepth, liu2023sparsebev} or equally distributed voxels~\cite{shi2020pv, shi2023pv++, rukhovich2022fcaf3d}, so that the advanced convolution networks can be directly applied. Pioneered by PointNet and its variants~\cite{qi2017pointnet, qi2017pointnet++}, point-based methods have become extensively employed on generating 3D object proposals directly from raw points~\cite{wang2022rbgnet, qi2019votenet, cheng2021back, xie2021venet, zhang2020h3dnet}. Most point-based methods can be considered as bottom-up approaches, which extract the point-wise features and group them to obtain object features. 
However, prior works focus on modifying network architectures while ignoring the distributional properties of point clouds, leading to inaccurate point assignments and sub-optimal grouping performances. Our proposed method exploits the distributional properties of point clouds and directly models the object center distribution using noise conditioned score network, which boosts 3D object detection performance.

\vspace{-2mm}
\subsection{Diffusion Models.}
\noindent As a class of probabilistic generative models, diffusion models have recently demonstrated remarkable results in fields including computer vision~\cite{ji2023ddp, gan2023v4d, shi2021mvnet}, natural language processing~\cite{chen2023maskdiffuionlm, he2022diffusionbert, lovelace2024latentdiffusionnlp}, multimodal data generation~\cite{gao2024guess, li2024pose, zhang2024text2nerf}, and temporal data modeling~\cite{bilovs2023modeling, yuan2023spatio}. 
Inspired by non-equilibrium thermodynamics, diffusion models learn to reverse a forward process that gradually degrades the training data structure and generates new data by recovering the data samples from initial samples in random distribution. 
Based on the forward and reverse process, diffusion models can be formulated as three sub-categories: Denoising Diffusion Probabilistic Models (DDPMs), Noise Conditioned Score Networks (NCSNs), and approaches that are based on stochastic differential equations and generalize over the first two sub-categories~\cite{croitoru2023diffusionsurvey}. 
Denoising diffusion probabilistic models slowly corrupt the training data using Gaussian noise until the data reaches Gaussian distribution then generate new samples from said Gaussian distribution by approximating the reverse steps.
On the other hand, noise conditioned score networks train a shared neural network via score matching to estimate the score function of the data distribution and then generate new samples by moving an initial sample to high-density regions based on the estimated score function of the data distribution. To accurately estimate the score function, noise conditioned score network first perturbs the data with Gaussian noise of different intensities and then jointly estimates the score functions of all noise-perturbed data distributions. 
While diffusion models have achieved great success in generative tasks, their potential for discriminative tasks has yet to be fully explored. In this work, we adopt noise conditioned score network to 3D object detection and achieve noticeable performance boost.

\vspace{-4mm}
\subsection{Diffusion Model for Point Clouds}
\noindent Since the success of diffusion models in the 2D image domain, researchers have started to explore the potential of diffusion models in various 3D point cloud tasks.
Luo~\cite{luo2021diffusion} is the first to use denoising diffusion probabilistic model for unconditional point cloud generation. They employ a point-wise network to generate point clouds, which is similar to a 2-stage PointNet.
Zhou~\cite{zhou2021ballqueryisbad} utilized a conditional diffusion model for point cloud completion by training a point-voxel convolution network. Furthermore, Vahdat~\cite{vahdat2022lion} proposed LION, a text-based 3D generation model to control the point cloud generation process by incorporating text-to-3d embedding into the diffusion model. Point-E~\cite{nichol2022pointe} first generates a single view image with GLIDE~\cite{nichol2021glide} and then performs image-to-shape diffusion. Wu~\cite{wu2023sketch} proposed a sketch and text guided probabilistic diffusion model for colored point cloud generation that conditions the denoising process jointly with a hand drawn sketch of the object and its textual description. Furthermore, DiffFacto~\cite{nakayama2023difffacto} models independent part style and part configuration distributions, and utilizes a novel cross diffusion network to generate coherent and plausible shapes. Lyu~\cite{lyu2021pdr} argued that point cloud completion can be treated as a conditional generation problem in the framework of diffusion models and proposed a novel Point Diffusion-Refinement (PDR) paradigm for point cloud completion. Luo~\cite{luo2021scoredenoise} adopted score matching for point cloud denoising. They proposed to increase the log-likelihood of each point from the noisy point cloud via gradient ascent.
Despite great success in these tasks, diffusion models are rarely explored for 3D object detection. Our proposed method exploits the distributional properties of point clouds and employs noise conditioned score network to model the distribution of object centers, which enables our model to generate more accurate votes and object proposals.

\begin{figure*}[t]
\centering
\includegraphics[width=\linewidth]{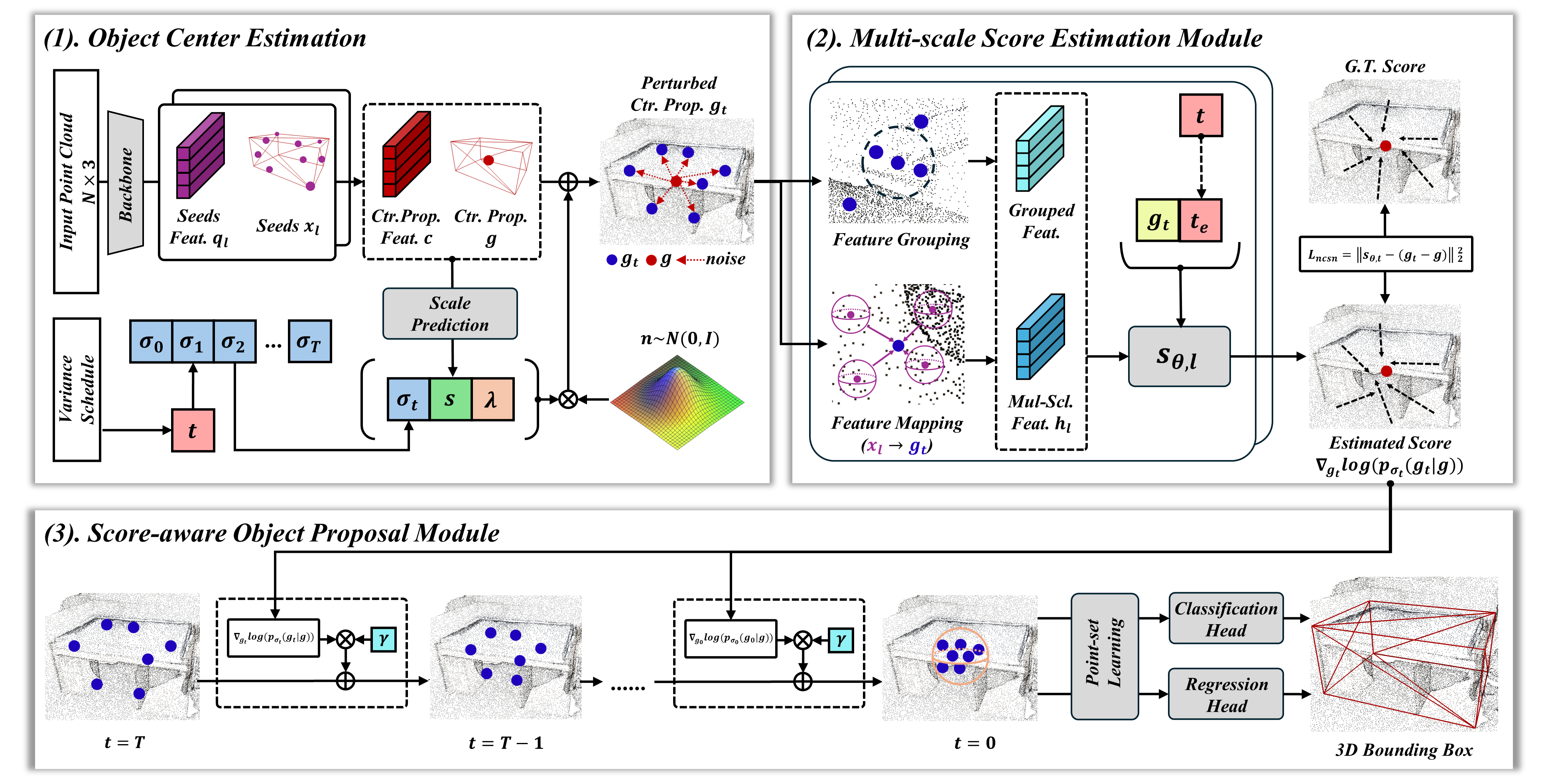}
\vspace{-0.7cm}
\caption{Overview of our proposed method. We first utilize a PointNet++ backbone to extract the point-wise features and generate a set of object center proposals. Random noises are then added to the generated proposals to corrupt the data. We propose a multi-scale score estimation module to predict the added random noise while conditioned on the input point cloud. Finally, we perform gradient ascent to denoise the perturbed object center proposals and generate 3D bounding boxes in our score-aware object proposal module.}
\label{fig_pipeline}
\vspace{-4mm}
\end{figure*}

\section{Methodology}
\subsection{Preliminaries}\label{section3.1}
\noindent \textbf{3D Object Detection: }Given a 3D point cloud scene $X=\left\{ x_i \right\}_{i=1}^N$ consisting of $N$ points, the goal of 3D object detection is to estimate oriented 3D bounding boxes and semantic labels of 3D objects in the scene. Specifically, we formulate the oriented 3D bounding boxes $b=\left\{ b_c,b_s,b_o \right\}$, where $b_c=\left\{ c_x,c_y,c_z \right\}$ denotes the center coordinate, $b_s=\left\{ d_x,d_y,d_z \right\}$ denotes the object size, and $b_o \in \mathbb{R}^1$ represents the bounding box orientation. 

\noindent \textbf{Noise Conditioned Score Networks: }
Noise Conditioned Score Networks (NCSNs)~\cite{song2019generative} are a class of probabilistic generative models that estimate the score function of data distribution and generate new samples by moving random samples toward samples in regions with high density, where the score function of data distribution $p(x)$ is defined as the gradient of the log density with respect to the input $\nabla_x\log_{}p(x)$.
Similar to Denoising Diffusion Probabilistic Models (DDPMs)~\cite{ho2020ddpm}, noise conditioned score networks perturb the training data with Gaussian noise of different intensities in the forward process and jointly estimate the score functions of all noise-perturbed data distributions during the reverse process. Formally, given a sequence of Gaussian noise scales $\sigma_1<\sigma_2<\dots<\sigma_T$ and an underlying data distribution $p(x)$, the forward process is defined as: 
\begin{equation}
    p_{\sigma_t}(x_t|x)=\mathcal{N}(x_t|x, \sigma_t^2\mathbf{I}),
\end{equation}
where $\mathcal{N}$ denotes Gaussian distribution and $\mathbf{I}$ denotes identity matrix. During training, a score network $s_\mathbf{\theta}$ is trained to estimate the score functions of each perturbed data distribution $p_{\sigma_t}(x_t|x)$. Specifically, the score network $s_\mathbf{\theta}$ is a neural network parameterized by $\boldsymbol\theta$, and is trained to estimate the score function of $p_{\sigma_t}(x_t|x)$ by minimizing the following objective:
\begin{equation}
    \frac{1}{T}\sum_{t=1}^{T}\lambda(\sigma_t)\mathbb{E}_{p(x)}\mathbb{E}_{p_{\sigma_t}(x_t|x)}\left \| s_{\theta}(x_t, \sigma_t)-\nabla_{x_t}\log_{}p_{\sigma_t}(x_t|x)  \right \|_2^2,
\end{equation}
where $\lambda(\sigma_t)$ represents weighting function. At inference time, new data samples are generated using annealed Langevin dynamics. Starting with an initial value $x_T \sim \pi(x)$ with $\pi$ being a prior distribution, annealed Langevin dynamics applies the following for a fixed number of iterations:
\begin{equation}
    x_{t-1} = x_{t} + \frac{\gamma }{2} \cdot \nabla_{x_t}\log_{}p_{\sigma_t}(x_t|x) + \sqrt\gamma \cdot z_t,
\end{equation}
where the score function $\nabla_{x_t}\log_{}p_{\sigma_t}(x_t|x)$ is given by the trained score network conditioned on the time step and $z_t \sim\mathcal{N}(0,\mathbf{I})$, $\gamma_t$ controls the magnitude of the update in the direction of the score function.

We tackle the 3D object detection task with a noise conditioned score network by moving from random points in a 3D point cloud toward the vicinity of object centers. To this end, we train our score network to estimate the score function of the distribution of the center coordinates $b_c$ of oriented 3D bounding boxes. 
As illustrated in Figure~\ref{fig_pipeline}, our method mainly consists of three parts: (1) object center estimation that models the center coordinate $b_c$ of bounding boxes as a set of samples from a 3D distribution and systematically adds random noises to the distribution(Section \ref{section3.2}), (2) multi-scale score estimation module that approximates the score function of the perturbed distribution conditioned on the point cloud $X$(Section \ref{section3.3}), and (3) score-aware object proposal module that moves 3D points in $X$ toward the high-density region of object centers and aggregates the scene context in the vicinity of object centers(Section \ref{section3.4}).

\subsection{Object Center Estimation}
\label{section3.2}
\noindent The forward process of noise conditioned score network involves adding Gaussian noise to data samples. We model the center of bounding boxes as a set of data samples from a distribution. However, the bounding box centers are unknown at inference time. Given the input point cloud $X$, we construct the forward process by first generating a set of object center proposals that lie close to ground truth object centers and then adding random noise to object center proposals.

\noindent \textbf{Object Center Proposal Generation. }
We employ the widely-used PointNet++~\cite{qi2017pointnet++} backbone as our backbone to extract point-wise features. Since object surface points play a pivotal role in predicting their associated object locations and orientations, we adopt the foreground sampling strategy used in RBGNet~\cite{wang2022rbgnet}. Specifically, we append a segmentation head after each set abstraction level in the backbone to score each point, which reflects its probability of belonging to the foreground. We then sort the segmentation score and select top-$k$ to form a foreground set and the rest are background set. The farthest point sampling is then applied to the foreground and background set separately to obtain the sub-sampled point set and point-wise features.

The PointNet++ backbone builds a hierarchical grouping of points and progressively abstracts larger and larger local regions along the hierarchy, which facilitates robust point set feature learning. To take advantage of deep features, we design a multi-scale object center proposal generation network. In particular, instead of only using the output of the final feature propagation level to predict the object center, we utilize the output of every feature propagation level to preserve point cloud details at different abstraction levels and generate accurate object center proposals. Given the point set $\left\{x_l \mid 1 \le l \le L \right\}$ and point-wise feature $\left\{q_l\mid 1 \le l \le L \right\}$ at $l$-th feature propagation level, the multi-scale object center proposal center generation network at $l$-th level outputs the Euclidean space offset $\Delta x_l$ and a feature offset $\Delta q_l$:
\begin{align}
\Delta x_l &= MLP(x_l, q_l), \\
\Delta q_l &= MLP(x_l, q_l).
\end{align}
The object center proposals $\Tilde{x}_l$ at $l$-th level and corresponding feature $\Tilde{q}_l$ are then calculated as $\Tilde{x}_l = x_l + \Delta x_l$ and $\Tilde{q}_l = q_l + \Delta q_l$ respectively. Our object center proposal generation network is supervised by the Chamfer Distance between the ground truth object centers and object center proposals during training:
\begin{equation}
    L_{ctr}=\frac{1}{L}\sum_{l=1}^L\left \| \Tilde{x}_l - b_c^{*}\right \| ,
\end{equation}
where $b_c^{*}$ represents ground truth object centers.
Finally, we concatenate all the center proposals $\left\{\Tilde{x}_l\mid 1 \le l \le L \right\}$ and apply the farthest point sampling to obtain the object center proposals $g$ and corresponding features $c$. 

\begin{figure}[!htbp]
\centering
\includegraphics[width=\linewidth]{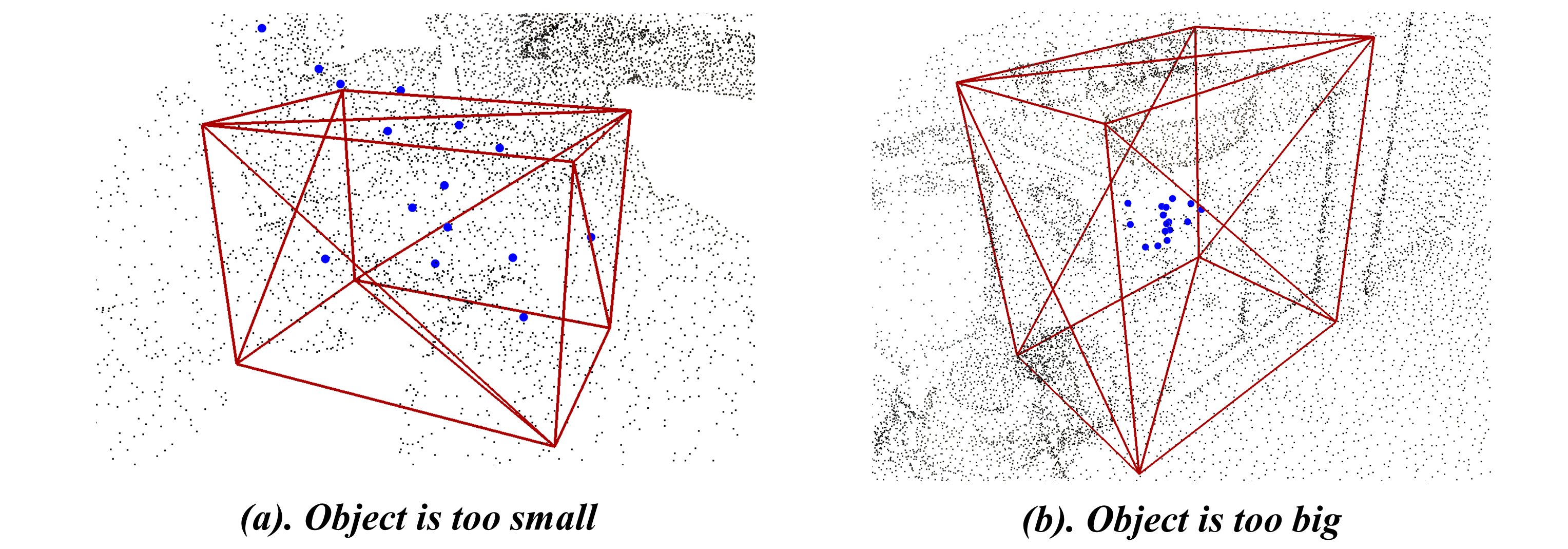}
\vspace{-0.7cm}
\caption{Instances of adding un-normalization noise. We can clearly see that the perturbed object center proposals in both instances fail to achieve uniform spatial coverage.}
\label{fig_norm}
\vspace{-4mm}
\end{figure}

\noindent \textbf{Object Center Corruption. }
The forward process of noise conditioned score network adds Gaussian noise to the data samples while the mean remains unchanged~\cite{song2019generative}. However, the inherent variations in 3D object sizes and shapes make this approach unfavorable. 
In practice, adding raw Gaussian noise to the center proposals of objects with different shapes and sizes can cause difficulties in generating high-quality samples. 
To illustrate the problem and demonstrate its negative effect, we provide visualization of two samples in Figure~\ref{fig_norm} where we add raw Gaussian noise to the center of two objects. As the figure demonstrates, the perturbed center proposals do not uniformly cover the overall shape of objects. In Figure~\ref{fig_norm}(a), most of the perturbed proposals are far from the bounding box since the object is too small, while the perturbed proposals in Figure~\ref{fig_norm}(b) have little coverage of the object bounding box as the bounding box is too large. With limited spatial coverage, noise conditioned score network is unable to accurately estimate the score function and aggregate scene context for object detection. To address this problem, we introduce a novel forward process that utilizes the object scale to enforce better spatial coverage.

Our forward process starts with predicting the object scale based on the object center proposals $g$ and their features $c$. Since the center proposals lie close to ground truth object centers, we utilize a set abstraction level to group and cluster the center proposals according to spatial proximity. Next, an object scale for each center proposal in a cluster is predicted by passing the cluster through a shared MLP:
\begin{equation}
    \Tilde{s}=MLP(SA(g, c)).
\end{equation}
We then calculate the half diagonal side of the estimated object scale:
\begin{equation}
    s = \frac{1}{2}\left \| \Tilde{s} \right \|_2 .
\end{equation}
Finally, our proposed forward process can be formulated as:
\begin{equation}
\label{forward}
    p_{\sigma_t}(g_t|g)=\mathcal{N}(g_t|g, s \lambda \sigma_t^2\mathbf{I}),
\end{equation}
where $s$ and $\lambda$ jointly control noise magnitudes. To be specific, $s$ normalizes the added noise of each center proposal by the corresponding object scale. $\lambda$ is a hyperparameter that controls the percentile of the perturbed proposals inside the bounding box. For Gaussian distribution, the values less than one standard deviation away from the mean account for $68.27\%$ of the set, therefore, we empirically set $\lambda=0.6$. 
In our settings, we repeat the corruption process for each object center proposal $16$ times, meaning each object center proposal corresponds to $16$ perturbed object center proposals.

\subsection{Multi-scale Score Estimation Module}
\label{section3.3}
\noindent Given the perturbed object center proposals $g_t$ and the input point cloud $X$, we aim to learn score estimates by predicting the noise added in the forward process. Since the perturbed center proposals uniformly cover the overall shape of objects, the score network should effectively incorporate fine-grained details and multi-scale information from the input point cloud $X$. 
To this end, we first extract the multi-scale features for each perturbed object center proposal while conditioned on the input point cloud $X$. 
We then design a multi-scale score network to jointly predict the noise and transform the point-wise features for subsequent object detection.

\noindent \textbf{Multi-Scale Feature Extraction. }
The input of our proposed multi-scale feature extraction is the perturbed object center proposals $g_t$ and the output of the PointNet++ backbone. For the output of the $l$-th feature propagation level, we first map the point-wise feature $\left\{q_l^i\mid 1\le i\le N_l \right\}$ at points $\left\{x_l^i\mid 1\le i\le N_l \right\}$ to the perturbed object center proposals $\left\{g_t^r\mid 1\le r\le N_g  \right\}$ using point grouping operation, where $N_l$ and $N_g$ are the number of sub-sampled points at level $l$ and the number of perturbed object center proposals. Previous works argue that ball query is only suitable for point clouds that have the same level of density as the input $X$, but can not handle point clouds close to Gaussian noise~\cite{zhou2021ballqueryisbad}. To address this problem, we find the K-nearest neighbors $\left\{ x_l^k \mid k\in \mathcal{B}(g_t^r)\right\} \in \left\{x_l^i\mid 1\le i\le N_l \right\}$ for each $g_t^r$ and map features $\left\{q_l^k\mid k\in \mathcal{B}(g_t^r) \right\}$ to $g_t^r$ (where $\mathcal{B}$ denotes K-nearest neighbor). In addition, we adopt subtraction as an anti-symmetric operation on the grouped points $\left\{ x_l^k \mid k\in \mathcal{B}(g_t^r)\right\}$ and $g_t^r$ to encode the local information $(x_l^k-g_t^r)$~\cite{wang2019dgcnn}.
We then concatenate the mapped features with the local information to obtain the initial features $\Tilde{h}_l^r$ for $g_t^r$ at level $l$:
\begin{equation}
    \Tilde{h}_l^r = [q_l^k; MLP((x_l^k-g_t^r))],
\end{equation}
where $[;]$ denotes concatenation. 
We then utilize a shared MLP and a trainable aggregation function to transform $\Tilde{h}_{l}^r$ into a single feature vector:
\begin{equation}
    h_l^r = \varphi(MLP(\Tilde{h}_l^r)),
\end{equation}
where $\varphi$ denotes the aggregation function and is realized with a non-linear transformation. 
Repeating this process for each feature propagation level in our backbone, we obtain the multi-scale features for each perturbed proposal: $\left\{ h_l \mid 1 \le l \le L\right\}$.

\noindent \textbf{Multi-scale Score Network. }
The goal of our proposed score estimation module is to predict the random noise added to the center proposals $g$. Since the multi-scale feature $h_l$ captures the geometry features of objects at different scales, we introduce a parallel estimation scheme to jointly estimate the score function at different levels. In particular, we design a score network $s_{\theta, l}$ for each level $l$ in the multi-scale features $h_l$ to predict the noise added in the forward process. Formally, the score network $s_{\theta, l}$ is trained to minimize the following objective:
\begin{equation}
\begin{split}
    \mathbf{L}_{ncsn}=\frac{1}{TL}\sum_{t=1}^{T}\sum_{l=1}^{L}&\lambda(\sigma_t)\mathbb{E}_{p(g)}\mathbb{E}_{p_{\sigma_t}(g_t|g)}, \\ 
    &\left \| s_{\theta, l}(g_t, t_e, h_l)-\nabla_{g_t}\log_{}p_{\sigma_t}(g_t|g)  \right \|_2^2,
\end{split}
\end{equation}
where $g_t$, $t_e$ and $h_l$ denote the coordinates of the perturbed object center proposal $g_t$, time step embedding, and multi-scale feature respectively. $s_{\theta, l}$ represents the score network at level $l$. Since $\nabla_{g_t}\log_{}p_{\sigma_t}(g_t|g)$ can be simply calculated as $(g_t-g)$, the above objective amounts to minimizing the following loss function:
\begin{equation}
\begin{split}
    \mathbf{L}_{ncsn}=\frac{1}{2\space TLB}\sum_{t=1}^{T}\sum_{l=1}^{L}\sum_{B}
    \left \| s_{\theta, l}(g_t, t_e, h_l)-(g_t-g)  \right \|_2^2,
\end{split}
\end{equation}
where $B$ and $|B|$ denote mini batch and mini batch size.

The score network $s_{\theta, l}$ at $l$-th level is mainly parameterized by the multi-scale feature $h_l$ of perturbed object center proposals $g_t$. In principle, we can solely use $h_l$ to predict the added noise. However, due to the randomness of the perturbed object center proposals, this approach may increase the bias of noise estimation and consequently negatively affect the performance. Therefore, we propose a grouping strategy to adaptively aggregate information and improve robustness according to the distributional properties of points. Specifically, we encode the local region feature of each perturbed object center proposal $g_t$ using a set abstraction level and concatenate with the multi-scale feature $h_l$:
\begin{equation}
    h_{l,e}=[h_l; SA(h_l, g_t)].
\end{equation}
The enhanced multi-scale feature $h_{l,e}$ not only summarizes the feature at the local region of each perturbed object center proposal but also provides additional information for noise estimation, which is beneficial for object proposal generation.
In addition, we also incorporate information of the time step embedding $t_e$, and the coordinate of the perturbed object center $g_t$ to the score network. Given time step $t$, we employ sinusoidal positional embedding to produce $t_e$ that consists of pairs of sines and cosines with varying frequencies, followed by two linear layers with LeakyReLU activation function.

The score network $s_{\theta, l}$ is realized through a series of linear transformations. Specifically, each linear transformation consists of three fully connected layers: transformation unit $\varphi_t$, gate unit $\varphi_g$, and bias unit $\varphi_b$. The calculation of the linear transformation can be formulated as:
\begin{align}
c_t &=\varphi_t(h_{l,e}), \\
c_g &= \mathcal{S}(\varphi_g([t_e; g_t])), \\
c_b &= \varphi_b([\sigma_t; g_t]), \\
d &= c_t^Tc_g + c_b,
\end{align}
where $\mathcal{S}$ denotes sigmoid function and $d$ is the output of the linear transformation. The last layer of the score network $s_{\theta, l}$ is a fully connected layer that takes the output from the last linear transformation as input and outputs the predicted noise $\Delta g_{t,l}$. We repeat this process for all levels in multi-scale feature $\left\{ h_l \mid 1 \le l \le L\right\}$ then aggregate the predicted noise using a mean-pooling layer to obtain the final predicted noise $\Delta g_t$.

\subsection{Score-aware Object Proposal Module}
\label{section3.4}
\noindent The Score-aware object proposal module is designed to generate object proposals by denoising and aggregating the perturbed object center proposals $g_t$. 
Equipped with the predicted noise $\Delta g_t$, we first denoise the object center proposals $g_t$ using gradient ascent instead of Langevin dynamics, as Langevin dynamics introduces unwanted randomness in the form of $z_t$.
\begin{equation}
\label{denoise}
    \Tilde{g} = g_t + \gamma\Delta g_t,
\end{equation}
where $\Tilde{g}$ denotes the denoised center proposals and $\gamma$ represents step size. 

The denoised center proposals $\Tilde{g}$ not only lie close to object centers but also contain fine-grained information of foreground objects in the form of multi-scale features $\left\{ h_l \mid 1 \le l \le L\right\}$. Therefore, we leverage a point set learning network to aggregate the denoised center proposals and generate high-quality object proposals. In particular, we treat the unperturbed object center proposals $g$ as query points and utilize K-nearest neighbor to find the $k$ nearest denoised center proposals $\left \{ \tilde{g}^j \mid j\in\mathcal{B}(g^i) \right\}$. 
We then retrieve the corresponding multi-scale feature $h_l^j$ and pass them through a PointNet-like module:
\begin{equation}
    h^j=MLP(\varphi(MLP(\left\{h_l^j\right\}_{l=1}^L)), \space\space\space j\in\mathcal{B}(g^i)
\end{equation}
where $\varphi$ denotes a non-linear transformation to aggregate the multi-scale feature $h_l^j$. The last part of the object proposal module consists of two parallel sub-networks for classification and regression tasks respectively. Taking $h^j$ as input, the classification and regression processes are formulated as:
\begin{align}
o_c &= MLP_c(\varphi(h^j)), \\
o_r &= MLP_r(\varphi(h^j)),
\end{align}
where $\varphi$ denotes a trainable aggregation operation over $\left\{ h^j \mid 1 \le j \le k\right\}$. We follow existing works~\cite{qi2019votenet} on how to parameterize the oriented 3D bounding boxes. Specifically, $o_r$ has $3+2N_H+4N_S$ channels where $N_H$ is the number of heading bins and $N_S$ is the number of size clusters. $o_c$ is a $2+N_C$-dimension vector with an objectness score and semantic classification scores, where $N_C$ is the number of semantic classes.

The loss function of our proposed model consists of object center estimation loss $\mathcal{L}_{ctr}$, denoising loss $\mathcal{L}_{ncsn}$, bounding box estimation loss $\mathcal{L}_{box}$, corner loss $\mathcal{L}_{corner}$, objectness loss $\mathcal{L}_{obj}$, and semantic classification $\mathcal{L}_{cls}$. Our final loss function can be formulated as:
\begin{equation}
\begin{split}
    \mathcal{L} =& w_{ctr}\mathcal{L}_{ctr} + w_{ncsn}\mathcal{L}_{ncsn} + w_{box}\mathcal{L}_{box} +\\ &w_{corner}\mathcal{L}_{corner} +
    w_{obj}\mathcal{L}_{obj} + w_{cls}\mathcal{L}_{cls}.
\end{split}
\end{equation}
The object center estimation loss $\mathcal{L}_{ctr}$ and denoising loss $\mathcal{L}_{ncsn}$ are used to supervise the object center proposal generation network and multi-scale score estimation module respectively. Following the settings used in VoteNet, we adopt the same formulation of bounding box estimation loss $\mathcal{L}_{box}$, objectness loss $\mathcal{L}_{obj}$, and semantic classification $\mathcal{L}_{cls}$. We additionally incorporate the corner loss~\cite{qi2018frustum} to jointly optimize the estimation of oriented bounding box. The bounding box corner loss $\mathcal{L}_{iou}$ is formulated as:
\begin{equation}
    \mathcal{L}_{corner} = \sum_{k=1}^8\left \| P_k-P_k^*\right \| ,
\end{equation}
where $P_k$ and $P_k^*$ denote the corners of the predicted bounding box and ground truth bounding box respectively.

\begin{algorithm}\scriptsize
\caption{Training}
\label{alg_train}
\begin{Verbatim}[commandchars=\\\{\}]
\textcolor{magenta}{def} train_loss(pc, gt_bbox, sigmas):
  \textcolor{Green}{# Extract points and features}
  bb_feat, pts = backbone(pc)

  \textcolor{Green}{# Generate object center proposals}
  p_ctr = pred_ctr(bb_feat)

  \textcolor{Green}{# Corrupt object center proposals }
  t = randint(1, T)
  ns = normal(mean=p_ctr, std=1)
  sigma_t = get_sigma(sigmas, t)
  p_scale = pred_scale(bb_feat)
  crpt_ctr = p_ctr + lambda * p_scale * sigma_t * ns

  \textcolor{Green}{# Denoise using multi-scale features }
  ms_feat = multi_scale_feat(bb_feat, crpt_ctr, pts)
  t_emb = time_emb(sigma_t)
  p_ns = denoise(ms_feat, t_emb, crpt_ctr)

  \textcolor{Green}{# Remove added noise and generate proposals }
  re_ctr = crpt_ctr + p_ns
  prp_bbox = prp_head(re_ctr, ms_feat)

  loss = get_loss(prp_bbox, gt_bbox)

  \textcolor{magenta}{return} loss
\end{Verbatim}
\end{algorithm}

\begin{algorithm}\scriptsize
\caption{Inference}
\label{alg_inference}
\begin{Verbatim}[commandchars=\\\{\}]
\textcolor{magenta}{def} inference(pc, sigmas, gammas, num_step):
  \textcolor{Green}{# Extract points and features}
  bb_feat, pts = backbone(pc)
  
  \textcolor{Green}{# Generate object center proposals}
  p_ctr = pred_ctr(bb_feat)

  \textcolor{Green}{# Start with random noise}
  t = T
  ns = normal(mean=p_ctr, std=1)
  sigma_t = get_sigma(sigmas, t)
  p_scale = pred_scale(bb_feat)
  crpt_ctr = p_ctr + lambda * p_scale * sigma_t * ns
  \textcolor{Green}{# Sampling starts}
  \textcolor{magenta}{for} t \textcolor{magenta}{in} sigmas:
    gamma_t = get_gamma(gammas, t)
    sigma_t = get_sigma(sigmas, t)

    ms_feat = multi_scale_feat(bb_feat, crpt_ctr, pts)
    t_emb = time_emb(sigma_t)
    p_ns = denoise(ms_feat, t_emb, crpt_ctr)
    \textcolor{Green}{# Gradient ascent}
    crpt_ctr = crpt_ctr + gamma_t * p_ns

  \textcolor{Green}{# Sampling ends}
  re_ctr = crpt_ctr
  prp_bbox = prp_head(re_ctr, ms_feat)

  \textcolor{magenta}{return} prp_bbox
\end{Verbatim}
\end{algorithm}

\subsection{Algorithm Pipeline}
\label{section3.5}
\noindent We devise a suitable training and inference procedure for generating object proposals using noise conditioned score network. During training, our proposed method learns to estimate the random noise added to the object center proposals and generates accurate object proposals by aggregating the denoised object center proposals. During inference, our proposed method generates denoised object center proposals from denoising random noise. 

\noindent \textbf{Training. }
Similar to the denoising diffusion probabilistic model, the training procedure of our proposed method involves two phases: the forward process and the backward denoising process. Given the input point cloud and ground truth object annotation, we generate the object center proposals and perturb them with normalized Gaussian noise to obtain the perturbed samples in the forward process. As in  (\ref{forward}), the Gaussian noise is controlled by object scale $s$, hyperparameter $\lambda$, and noise scale $\sigma_t$. Given the noise scales $\sigma_1<\sigma_2<\dots<\sigma_T$, we randomly select one noise scale for each batch during training. 
Subsequently, the backward denoising process is represented by reversing the forward process. We extract the multi-scale feature of each perturbed object center proposal and utilize our proposed denoising network to predict the added Gaussian noise. Finally, we employ our score-aware object proposal module to move perturbed center proposals toward the original object center proposals. During training, we run gradient ascent only once and set the step size $\gamma=1$. The final object proposals are then generated by aggregating the denoised object center proposals.

\noindent \textbf{Inference. }
Since the object center is unknown during inference, our inference procedure starts with generating the object center proposals. 
During inference, our proposed method generates denoised object center proposals by iterative sampling from perturbed data distributions using gradient ascent. 
We start the sampling process by initializing the samples $g_T\sim p_{\sigma_{T}}(g_{T})$, then run gradient ascent described in (\ref{denoise}) to move from the initial samples $g_T$ toward $g_{T-1}$ with step size $\gamma_{T}$, which equals to sample from $p_{\sigma_{T-1}}(g_{T-1})$. We then run gradient ascent to move from $g_{T-1}$ toward $g_{T-2}$ starting from the final samples of the previous simulation and using a reduced step size $\gamma_{T-1}$. We repeat this process, using the final samples of gradient ascent for $p_{\sigma_{t}}(g_{t})$ as the initial samples of gradient ascent for $p_{\sigma_{t-1}}(g_{t-1})$ and tuning down the step size $gamma_t$ gradually. Finally, we run gradient ascent to move $g_{1}$ toward $g_{0}$, which is close to the original object center proposals. To take full advantage of the denoising process, instead of only using the multi-scale feature of $g_1$, we calculate the mean multi-scale feature of all $g_t$ in the denoising trajectory.

\section{Experiments}
\subsection{Experimental Configuration}
\noindent We evaluate our model on two large-scale indoor 3D scene datasets, i.e., ScanNet V2~\cite{dai2017scannet} and SUN RGB-D~\cite{song2015sunrbgd}, and follow the standard data splits for both. We adopt the standard mean average precision (mAP) under different IoU thresholds (0.25 and 0.5) as the evaluation metric.

SUN RGB-D is a single-view dataset containing 10,335 indoor RGB-D images with annotated object bounding boxes and semantic labels. Following the data pre-processing in VoteNet, we convert each RGB-D image to point cloud by projecting the depth information onto the RGB image using the provided parameters. We consider the 10 most common object categories for evaluation, similar to prior works.

ScanNet V2 is a popular indoor scene dataset comprising 1513 3D reconstructions of indoor scenes out of which 1201 are reserved for training, and the remaining 312 are left for validation. Each scene contains axis-aligned 3D bounding box annotations for 18 object categories. We sample point clouds from the reconstructed meshes following VoteNet.

Our model is implemented in Pytorch, and trained using four NVIDIA RTX 3090 for 64 epochs with the ADAM optimizer. We use an initial learning rate of 0.001, weight decay of 0.5. The gradnorm clip is applied to stabilize the training dynamics.

\begin{table}[t]
\caption{Performance comparison with state-of-the-art on the SUN RGB-D val set. The number within the bracket is the average of 5 trials and the one outside is the best result. \label{tab_comp_sunrgbd}}
\vspace{-3mm}
\renewcommand{\arraystretch}{1.3}
\setlength{\tabcolsep}{5pt}
\centering
\begin{tabular}{lccccc}
\hline
Method & Backbone & mAP@0.25 & mAP@0.50 \\
\hline
VoteNet~\cite{qi2019votenet} & PointNet++ & 59.1 & 35.8  \\
HGNet~\cite{chen2020hgnet} & GU-Net & 61.6 & -  \\
MLCVNet~\cite{xie2020mlcvnet} & PointNet++ & 59.8 & -  \\
H3DNet~\cite{zhang2020h3dnet} & 4$\times$PointNet++ & 60.1 & 39.1 \\
BRNet~\cite{cheng2021back} & PointNet++ & 61.1 & 43.7 \\
3DETR~\cite{misra20213detr} & PointNet++ & 59.1 & 32.7 \\
VENet~\cite{xie2021venet} & PointNet++ & 62.5 & 39.2 \\
Group-free~\cite{liu2021groupfree} & PointNet++ & 63.0 (62.6) & 45.2 (44.4)  \\
RBGNet~\cite{wang2022rbgnet} & PointNet++ & 64.1 (63.6) & 47.2 (46.3)  \\
Ours & PointNet++ & \textbf{65.2 (64.4)} & \textbf{49.1 (48.5)}  \\
\hline
\end{tabular}
\end{table}

\begin{table}[t]
\caption{Performance comparison  with state-of-the-art on the ScanNet V2 val set. PointNet++w2× denotes the backbone width is expanded by 2 times. The number within the brackets is the average of 5 trials and the one outside is the best result. \label{tab_comp_scannet}}
\vspace{-3mm}
\renewcommand{\arraystretch}{1.3}
\setlength{\tabcolsep}{5pt}
\centering
\begin{tabular}{lccccc}
\hline
Method & Backbone & mAP@0.25 & mAP@0.50 \\
\hline
VoteNet~\cite{qi2019votenet} & PointNet++ & 62.9 & 39.9 \\
MLCVNet~\cite{xie2020mlcvnet} & PointNet++ & 64.5 & 41.4 \\
HGNet~\cite{chen2020hgnet} & GU-Net & 61.3 & 34.4 \\
3D-MPA~\cite{engelmann20203dmpa} & MinkNet & 64.2 & 49.2 \\
H3DNet~\cite{zhang2020h3dnet} & 4$\times$PointNet++ & 64.4 & 43.4 \\
VENet~\cite{xie2021venet} & PointNet++ & 67.7 & - \\
3DETR~\cite{misra20213detr} & PointNet++ & 65.0 & 47.0 \\
BRNet~\cite{cheng2021back} & PointNet++ & 66.1 & 50.9 \\
Group-free~\cite{cheng2021back} & PointNet++ & 67.2 (66.6) & 49.7 (49.0) \\
RBGNet~\cite{wang2022rbgnet} & PointNet++ & 70.2 (69.6) & 54.2 (53.6) \\
Ours & PointNet++ & \textbf{71.1 (70.2)} & \textbf{54.8 (53.9)} \\
\hline
Group-free~\cite{liu2021groupfree} & PointNet++w2$\times$ & 69.1 (68.6) & 52.8 (51.8) \\
RBGNet~\cite{wang2022rbgnet} & PointNet++w2$\times$ & 70.6 (69.9) & 55.2 (54.7) \\
Ours & PointNet++w2$\times$ & \textbf{71.6 (70.9)} & \textbf{56.0 (55.2)} \\
\hline
\end{tabular}
\end{table}

\subsection{Comparison with State-of-the-Art}
\noindent We compare our model with a wide range of state-of-the-art point-based 3D object detection models on ScanNet V2 and SUN RGB-D datasets. We report the best and average results to reduce randomness.

\noindent \textbf{SUN RGB-D. }
We first evaluate our model against several competing approaches on SUN RGB-D dataset. The results are shown in Table~\ref{tab_comp_sunrgbd}. Our model achieves $65.2\%$ and $49.1\%$ on mAP@0.25 and mAP@0.5, outperforming VoteNet~\cite{qi2019votenet} by $6.1\%$ on mAP@0.25 and $13.3\%$ on mAP@0.5, This is mainly because our model generates significantly more accurate votes than VoteNet. More importantly, our model outperforms other voting-based models, such as MLCVNet~\cite{xie2020mlcvnet}, VENet~\cite{xie2021venet}, BRNet~\cite{cheng2021back}, and RBGNet~\cite{wang2022rbgnet}. Existing voting-based models incorporate multi-level contextual learning, back-tracing, and self-attention based transformer to the voting process. While proven effective, they ignore the distributional properties of point clouds and fail to generate accurate votes when the input point cloud is noisy.
Benefiting from the impressive generative capability of noise conditioned score network, our model is capable of generating high-quality votes when dealing with noisy and incomplete point clouds as noise conditioned score network directly models the distribution of object centers, leading to better generalization capacity and detection performance.
Lastly, our model surpasses other approaches such as H3DNet~\cite{zhang2020h3dnet} and 3DETR~\cite{misra20213detr}, suggesting that our proposed 3D object detection paradigm is a better approach to 3D object detection.

\noindent \textbf{ScanNet V2. }
We also report the results on ScanNet V2 dataset, which contains a larger number of complete 3D reconstructed meshes as well as object categories. As shown in Table~\ref{tab_comp_scannet}, our model achieves $71.1\%$ mAP@0.25 and $54.8\%$ mAP@0.5 with one PointNet++ backbone. Our model outperforms H3DNet~\cite{zhang2020h3dnet} which uses 4 PointNet++ backbones with a significant margin. Owing to the iterative sampling process of noise conditioned score network, our model can fully exploit object surface geometry and generate high-quality object proposals using only one PointNet++ backbone.
Compared to RBGNet~\cite{wang2022rbgnet}, our model improves the mAP@0.25 by $1.0\%$ and mAP@0.5 by $0.8\%$ when using PointNet++w2$\times$ backbone. RBGNet utilizes a group of pre-determined rays uniformly emitted from vote cluster centers and then generates anchor points sampled on each ray to extract the local feature of the point cloud. Although this approach is effective, it generates redundant anchor points for each object proposal, which is both time and space-consuming. Our model employs diffusion model to generate a much smaller set of perturbed object center proposals to encode the object surface geometry.
However, we notice that the overall improvement is not as significant as on SUN RGB-D. 
This is mainly because most 3D scans in ScanNet V2 have more uniform point sampling and less noisy points than SUN RGB-D, making noisy votes a less prominent problem.

\subsection{Ablation Study}
We conduct extensive ablation studies on the val set of SUN RGB-D to analyze individual modules of our proposed model.

\begin{table}[t]
\caption{Quantitative results of different backbone network configurations\label{tab_abl_1_1}}
\vspace{-3mm}
\renewcommand{\arraystretch}{1.3}
\setlength{\tabcolsep}{10pt}
\centering
\begin{tabular}{lcc}
\hline
Method & mAP@0.25 & mAP@0.50 \\
\hline
w/o Multi-scale & 63.1 & 46.7 \\
w/ Pooling & 63.8 & 48.4 \\
w/o $\lambda$ and $s$ & 59.5 & 38.2 \\
w/o $\lambda$ & 61.7 & 44.8 \\
w/o $s$ & 62.4 & 45.1 \\
\hline
Ours & \textbf{65.2} & \textbf{49.1}  \\
\hline
\end{tabular}
\end{table}

\noindent \textbf{Object Center Estimation. }
We first examine the object center estimation module to learn about its impact on the rest of the model in Table~\ref{tab_abl_1_1} and ~\ref{tab_abl_1_4}. 
Table~\ref{tab_abl_1_1} shows that dropping the output of the feature propagation level other than the final level decreases the mAP@0.25 and mAP@0.5 by $2.1\%$ and $2.4\%$ respectively. This drop in performance indicates that point cloud details at different abstraction levels play a crucial role in generating accurate object center proposals. 
We can also see that applying a pooling operation to the output of all feature propagation levels before generating object center proposals slightly decreases performance (2nd row). We argue that applying pooling operations, such as max-pooling and mean-pooling, may lose some information about the point cloud at specific abstraction levels.

To analyze the importance of our proposed object center corruption process, we conduct several experiments using different formulations of the forward process. As shown in Table~\ref{tab_abl_1_1}, dropping both the predicted object scale $s$ and normalization term $\lambda$ decreases the performance by a significant margin of $5.7\%$ and $10.9\%$ respectively. In addition, dropping either $s$ or $\lambda$ also leads to sub-optimal performance. Due to the different shapes and sizes of objects, it is difficult to achieve uniform spatial coverage without normalization.

\begin{table}[t]
\caption{Results for different numbers of perturbed object center
proposals per center proposal. \label{tab_abl_1_4}}
\vspace{-3mm}
\renewcommand{\arraystretch}{1.3}
\setlength{\tabcolsep}{10pt}
\centering
\begin{tabular}{ccc}
\hline
Number of $g_t$ & mAP@0.25 & mAP@0.50 \\
\hline
4 & 63.6 & 47.4 \\
8 & 64.1 & 48.7 \\
16 & \textbf{65.2} & \textbf{49.1}  \\
22 & 63.3 & 47.8 \\
32 & 63.2 & 47.6 \\
\hline
\end{tabular}
\end{table}
Our object center corruption process also works well for a range of hyperparameters. 
Table~\ref{tab_abl_1_4} ablates the impact of various number of perturbed center proposals. We can see that our model performs best when generating $16$ perturbed center proposals per center proposal. We also find that generating more than 22 perturbed proposals hurts performance by about $2\%$.

\begin{table}[t]
\caption{Results for different configurations of the multi-scale score estimation module. \label{tab_abl_2_1}}
\vspace{-3mm}
\renewcommand{\arraystretch}{1.3}
\setlength{\tabcolsep}{10pt}
\centering
\begin{tabular}{lcc}
\hline
Method & mAP@0.25 & mAP@0.50 \\
\hline
ball query & 62.6 & 46.3 \\
ball query+position & 64.9 & 49.0 \\
w/o Grouping & 64.1 & 48.3 \\
\hline
Ours & \textbf{65.2} & \textbf{49.1}  \\
\hline
\end{tabular}
\end{table}

\noindent \textbf{Multi-scale Score Estimation Module. }
We analyze the effectiveness of our multi-scale score estimation module. First, we examine the multi-scale feature extraction module. As shown in Table~\ref{tab_abl_2_1}, replacing the K-nearest neighbors with ball query decreases the performance by a margin of $2.6\%$ and $2.8\%$ on mAP@0.25 and mAP@0.5. This suggests that ball query is unable to find enough neighbors for each perturbed object center proposal. No information can be extracted when points do not have neighbors. K-nearest neighbor guarantees that a point always has $K$ neighbors so that meaningful information of the local point cloud can be extracted via MLP and aggregation.
Moreover, similar to Lyu~\cite{zhou2021ballqueryisbad}, attaching the absolute position of each perturbed center proposal to its feature before applying ball query achieves better results.

We also validate the potential of our multi-scale score network in predicting the added random noise. The results are shown in Table~\ref{tab_abl_2_1}. We can see that removing the grouping strategy in our score network decreases performance by about $1.1\%$ and $0.8\%$ on mAP@0.25 and mAP@0.5. This drop in performance indicates that incorporating information from adjacent center proposals is vital for predicting added noise and improving robustness. 

\begin{table}[t]
\caption{Results for different number of iteration steps.\label{tab_abl_3_1}}
\vspace{-3mm}
\renewcommand{\arraystretch}{1.3}
\setlength{\tabcolsep}{10pt}
\centering
\begin{tabular}{ccc}
\hline
Number of iteration & mAP@0.25 & mAP@0.50 \\
\hline
1 & 64.1 & 48.8 \\
2 & 64.2 & 48.8 \\
5 & 64.5 & 49.0 \\
10 & \textbf{65.2} & \textbf{49.1}  \\
20 & 64.6 & 49.0 \\
30 & 64.5 & 48.7 \\
\hline
\end{tabular}
\end{table}

\begin{table}[t]
\caption{Results for different phases of the denoising process.\label{tab_abl_3_2}}
\vspace{-3mm}
\renewcommand{\arraystretch}{1.3}
\setlength{\tabcolsep}{10pt}
\centering
\begin{tabular}{lcc}
\hline
Phases & mAP@0.25 & mAP@0.50 \\
\hline
Last & 63.9 & 48.5 \\
First Half & 64.4 & 48.8 \\
Second Half & 64.1 & 48.6  \\
Full & \textbf{65.2} & \textbf{49.1}  \\
\hline
\end{tabular}
\end{table}

\noindent \textbf{Score-aware Object Proposal Module. }
Finally, we verify the effectiveness of our proposed score-aware object proposal module. Since the iterative denoising process in our proposal module plays an essential role in aggregating scene context and predicting 3D bounding boxes, we conduct experiments to understand how iterative denoising works.
We first investigate the impact of different iteration steps. Table~\ref{tab_abl_3_1} shows that our model works well in a wide range of iteration steps. Iterating more steps improves the mAP@0.25 by up to $1.1\%$. However, our model performs worse when iterating more than 10 steps. This is mainly because more iterations inevitably include irrelevant information about the underlying objects and negatively impact the detection performance. 

We also investigate the effectiveness of different phases of the denoising process by using the multi-scale feature of different denoised $g_t$. 
As shown in Table~\ref{tab_abl_3_2}, using the multi-scale feature of all intermediate $g_t$ yields the best results. We can also see that using the first half the denoised $g_t$ performs better than the second half, suggesting that the multi-scale feature of $g_t$ during the initial denoising process provides richer information about the object surface geometry.

\subsection{Further Analysis}
\begin{figure}[!tbp]
\centering
\includegraphics[width=\linewidth]{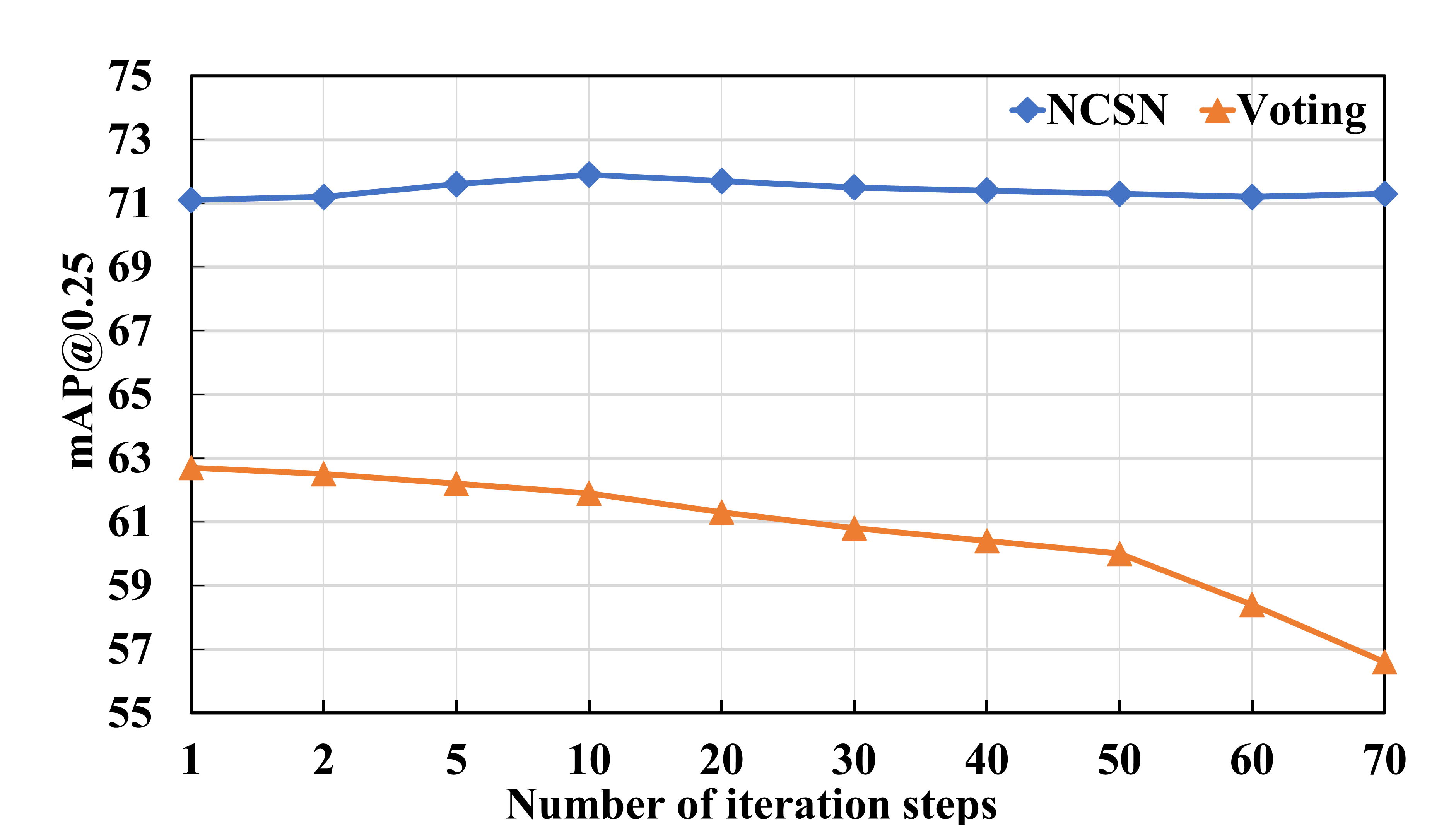}
\vspace{-7mm}
\caption{Comparing voting mechanism (Voting) to the proposed noise conditioned score network (NCSN) on the ScanNet V2 val set.}
\label{fig_vs_voting}
\vspace{-4mm}
\end{figure}

\noindent \textbf{Analysis on the NCSN. }
We compare the noise conditioned score network in our proposed model to the voting mechanism to show the effectiveness of the former. Since our proposed method utilizes multi-scale features and iterative denoising, we also incorporate multi-scale features and iterative voting to VoteNet~\cite{qi2019votenet} for a fair comparison. The results are shown in Figure ~\ref{fig_vs_voting}.

Our results in Figure~\ref{fig_vs_voting} lead to a few key observations: (1) Noise conditioned score network with 10 iteration steps achieves the best performance while voting mechanism performs the best using one iteration step. (2) As the iteration steps increase, the performance of noise conditioned score network rises while the performance of voting mechanism drops. (3) The best performance of noise conditioned score network is higher than that of voting mechanism. 
(4) The performance of our noise conditioned  score network stays relatively more stable with increasing iterations whereas the performance of voting mechanism drops significantly. 
These results indicate that the noise conditioned score network in our model is able to accurately estimate the score function of object center distribution at any location. Moreover, benefiting from the noise conditioned score network and gradient ascent, our model refines the data sample in a progressive way. In comparison, previous voting-based methods generate votes in a single step without refinement. These approaches not only perform worse when using two or more iteration steps but also lack the generalization capability of noise conditioned score network.

\begin{figure}[!tbp]
\centering
\includegraphics[width=\linewidth]{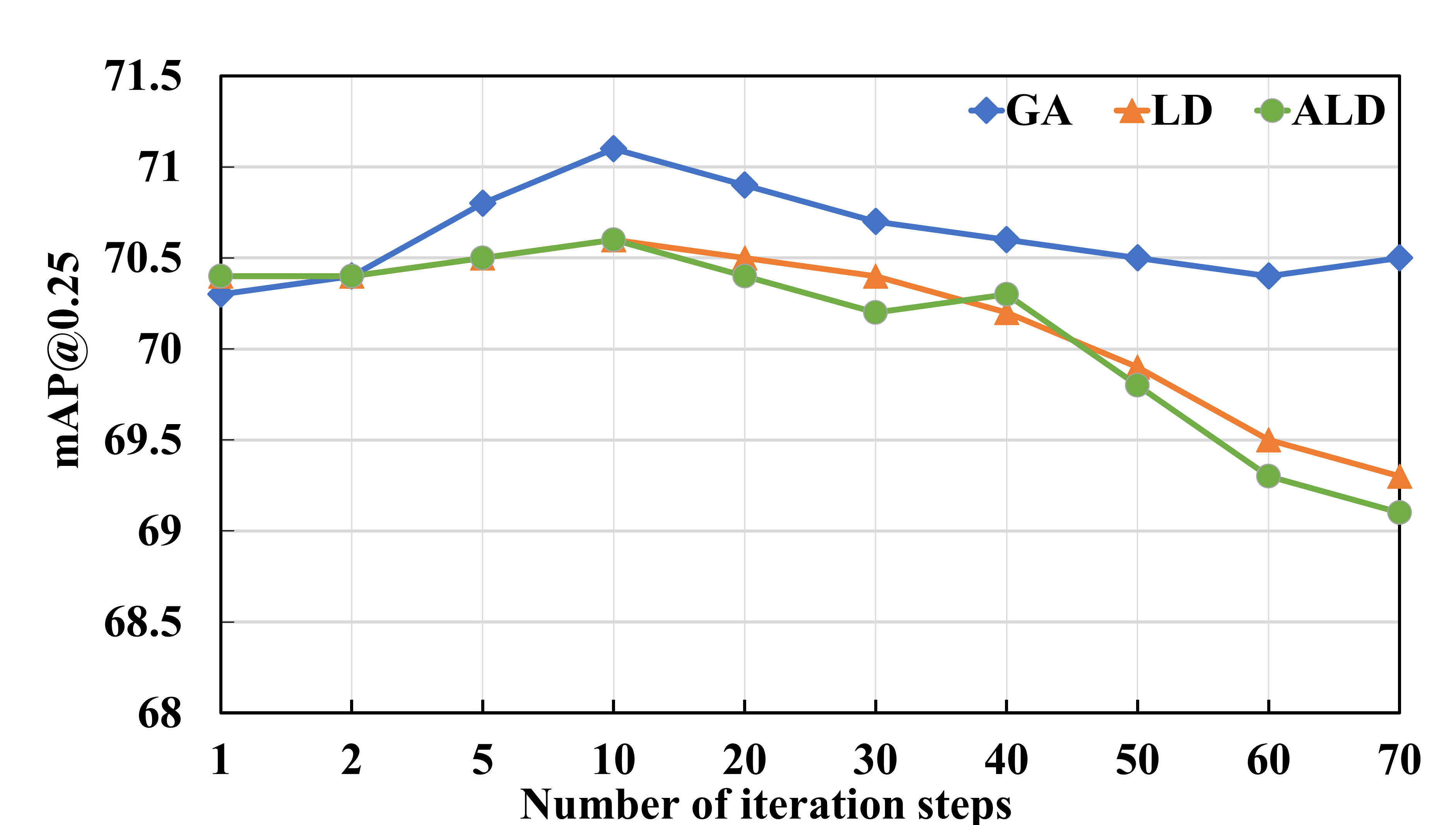}
\vspace{-7mm}
\caption{Performance comparison between Gradient Ascent (GA), Langevin Dynamics (LD), and annealed Langevin Dynamics (ALD) on the ScanNet V2 val set.}
\label{fig_vs_ld_ald}
\vspace{-1mm}
\end{figure}

\noindent \textbf{Analysis of Gradient Ascent. }
To validate the effectiveness of gradient ascent during the denoising process, we conduct additional experiments to compare its performance against Langevin dynamics and annealed Langevin dynamics. To be fair, we adopt the same step size for all three. The results are reported in Figure~\ref{fig_vs_ld_ald}. 
We can clearly see that gradient ascent outperforms Langevin dynamics and annealed Langevin dynamics on most iteration steps. In addition, as the iteration steps increase, the performance of Langevin dynamics and annealed Langevin dynamics decreases drastically while gradient ascent experiences a relatively smaller drop.
Although Langevin dynamics and annealed Langevin dynamics can produce samples from a target distribution based on its score function, they introduce randomness to the sampling process in the form of random Gaussian noise $z_t \sim \mathcal{N}(0, \textbf{I})$. In our proposed method, random noise $z_t$ inevitably causes displacement of the object center proposals, leading to inaccurate denoised results. Moreover, the displacement becomes more noticeable as the iteration step increases. In contrast, gradient ascent moves samples to high-density regions along the estimated vector field of the score function without randomness, achieving better performance.

\noindent \textbf{Comparison with DDPM. }
Denoising Diffusion Probabilistic Model~\cite{ho2020ddpm} has shown success in many point cloud related tasks. Therefore, we conduct experiments to investigate its impact on our detection pipeline. Specifically, we replace the noise conditioned score network in our model with a denoising diffusion probabilistic model to perturb and denoise the object center proposals.
Since the forward process of denoising diffusion probabilistic model is formulated as $q(x_t|x_0)=\mathcal{N}(x_t;\sqrt{\bar{\alpha}_t},(1-\bar{\alpha}_t)\mathbf{I})$, we evaluate the performance of denoising diffusion probabilistic model using a variety of variance schedules $\beta_1,...,\beta_T$ to find the best result. We also apply the same normalization (\textit{i.e.} object scale $s$ and hyperparameter $\lambda$) for a fair comparison.

As shown in Table~\ref{tab_ana_2}, even with DDIM~\cite{song2020ddim} sampling, denoising diffusion probabilistic model still performs worse than our noise conditioned score network. This is because the denoising diffusion probabilistic model destroys information in data and replaces it with Gaussian noise, making it impossible for perturbed center proposals to cover the 3D bounding box, leading to poor sampling quality and sub-optimal performance. The proposed noise conditioned score network adds Gaussian noise with zero mean to the data, which not only preserves the initial point cloud structure but also facilitates efficient sampling process.

\begin{figure*}[!htbp]
\centering
\includegraphics[width=\linewidth]{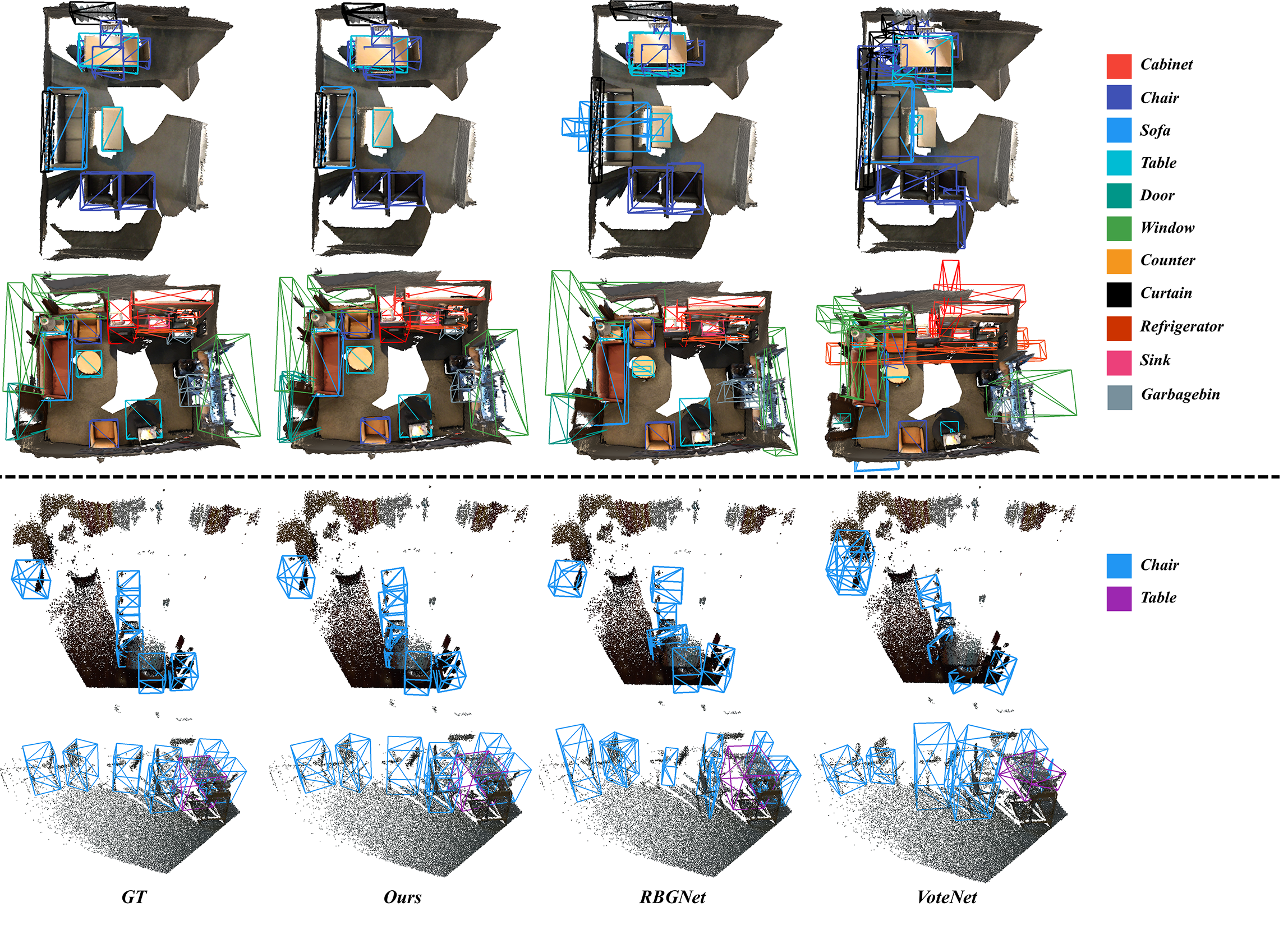}
\vspace{-9mm}
\caption{Qualitative results on ScanNet V2 (top) and SUNRGB-D (bottom) datasets. The baseline methods are VoteNet and RBGNet.}
\label{fig_viz_example}
\vspace{-4mm}
\end{figure*}

\begin{table}[t]
\caption{Comparison with Denoising Diffusion Probabilistic Model (DDPM)\label{tab_ana_2} on ScanNet V2}
\vspace{-3mm}
\renewcommand{\arraystretch}{1.3}
\setlength{\tabcolsep}{10pt}
\centering
\begin{tabular}{lcc}
\hline
Method & mAP@0.25 & mAP@0.50 \\
\hline
DDPM & 68.6 & 53.2 \\
DDPM w/ DDIM~\cite{song2020ddim} & 68.7 & 53.3 \\
\hline
Ours(NCSN) & \textbf{71.1} & \textbf{54.8}  \\
\hline
\end{tabular}
\end{table}

\subsection{Qualitative Results}
\noindent Figure~\ref{fig_viz_example} visualizes the detection results using VoteNet, RBGNet, and our proposed model on ScanNet V2 and SUNRGB-D. We can see that our proposed model can handle a wide range of indoor scenes and object categories, including \textit{Cabinet}, \textit{Chair}, \textit{Sofa}, \textit{Table}, and \textit{Window}. From the top of Figure~\ref{fig_viz_example}, We can observe that our proposed model successfully detects the overlapping objects and obtains their clean 3D bounding boxes. Our model also performs better when the input point cloud is incomplete and objects are sparse (see the bottom of Figure~\ref{fig_viz_example}). 

\begin{figure*}[!htbp]
\centering
\includegraphics[width=\linewidth]{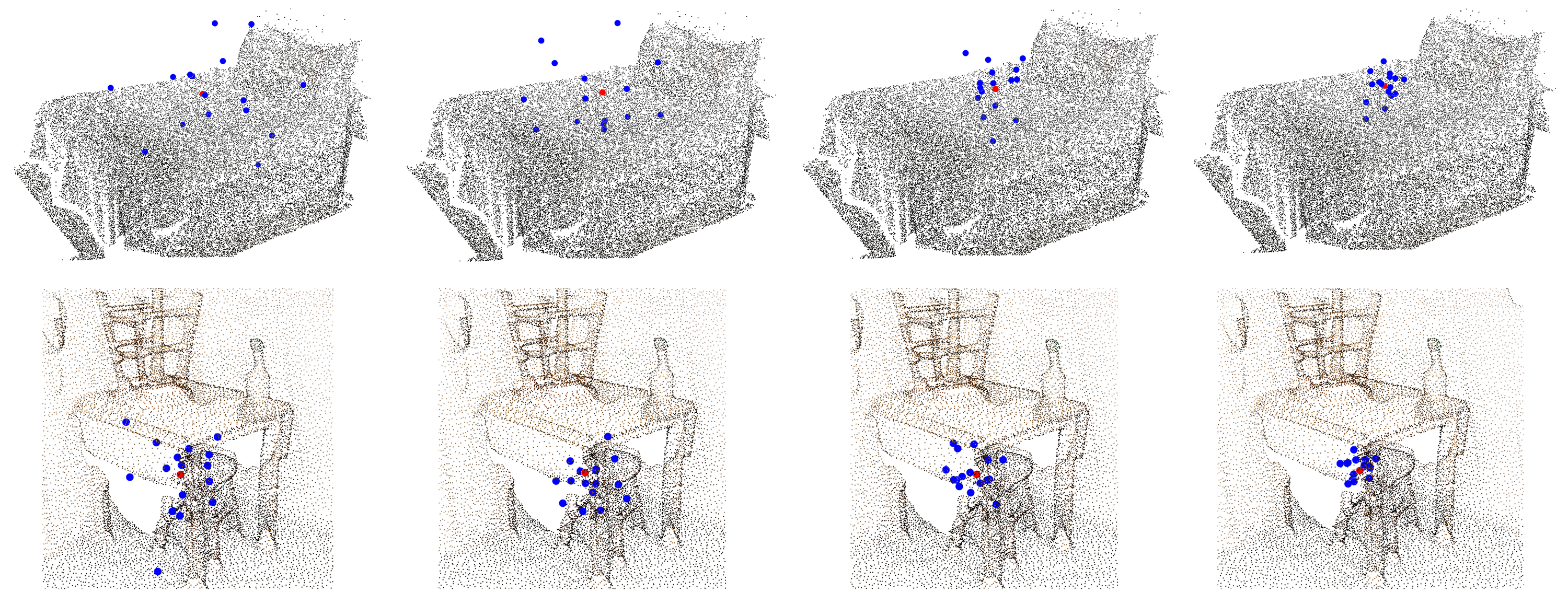}
\vspace{-5mm}
\caption{visualization of the denoising process of our proposed model, blue and red dots represent perturbed object center proposal $g_t$ and initial object center proposal $g$ respectively. }
\label{fig_viz_steps}
\end{figure*}

To demonstrate the effectiveness of our noise conditioned score network, we additionally visualize the the gradient ascent process of our model in Figure~\ref{fig_viz_steps}. We can observe that the perturbed object center proposals gradually converge to the vicinity of the object centers, indicating that the predicted score function is accurate and the gradient ascent process is effective at removing the added noise.

\begin{table}[t]
\caption{Performance comparison with existing point-based semantic segmentation models. \label{tab_semseg}}
\vspace{-3mm}
\renewcommand{\arraystretch}{1.3}
\setlength{\tabcolsep}{10pt}
\centering
\begin{tabular}{lcc}
\hline
Method & Val mIoU & Test mIoU \\
\hline
PointNet++~\cite{qi2017pointnet++} & 53.5 & 33.9 \\
PointCNN~\cite{li2018pointcnn} & - & 45.8  \\
PointConv~\cite{wu2019pointconv} & 61.0 & 55.6 \\
KpConv~\cite{thomas2019kpconv} & 69.2 & 68.0 \\
RPNet~\cite{ran2021rpnet} & - & 68.2 \\
\hline
Ours & \textbf{70.7} & \textbf{69.5} \\
\hline
\end{tabular}
\end{table}

\subsection{Extending to Semantic Segmentation Task}
One of the benefits of noise conditioned score network is the ability to move random 3D points toward the high-density region of a distribution. To further demonstrate the effectiveness of our noise conditioned score network and the estimated score function, we adopt our model for semantic segmentation in indoor point clouds. 

In our proposed method, we perform 3D object detection by moving random 3D points toward the vicinity of object centers to generate 3D bounding boxes and predict the corresponding semantic labels. As the goal of point cloud semantic segmentation task is to assign a category label to each discrete point in the input point clouds, we extend from our model by moving points in the input point cloud toward the object centers and predict the per point semantic labels instead. More specifically, we train noise conditioned score network to estimate the score function of the object center distribution by perturbing the generated object center proposals. We then predict the semantic category of denoised center proposals by aggregating the multi-scale feature. During inference, instead of moving random 3D points, we move the points in the input point cloud to their corresponding object centers and predict the semantic category of each point.  

\begin{figure}[!tbp]
\centering
\includegraphics[width=\linewidth]{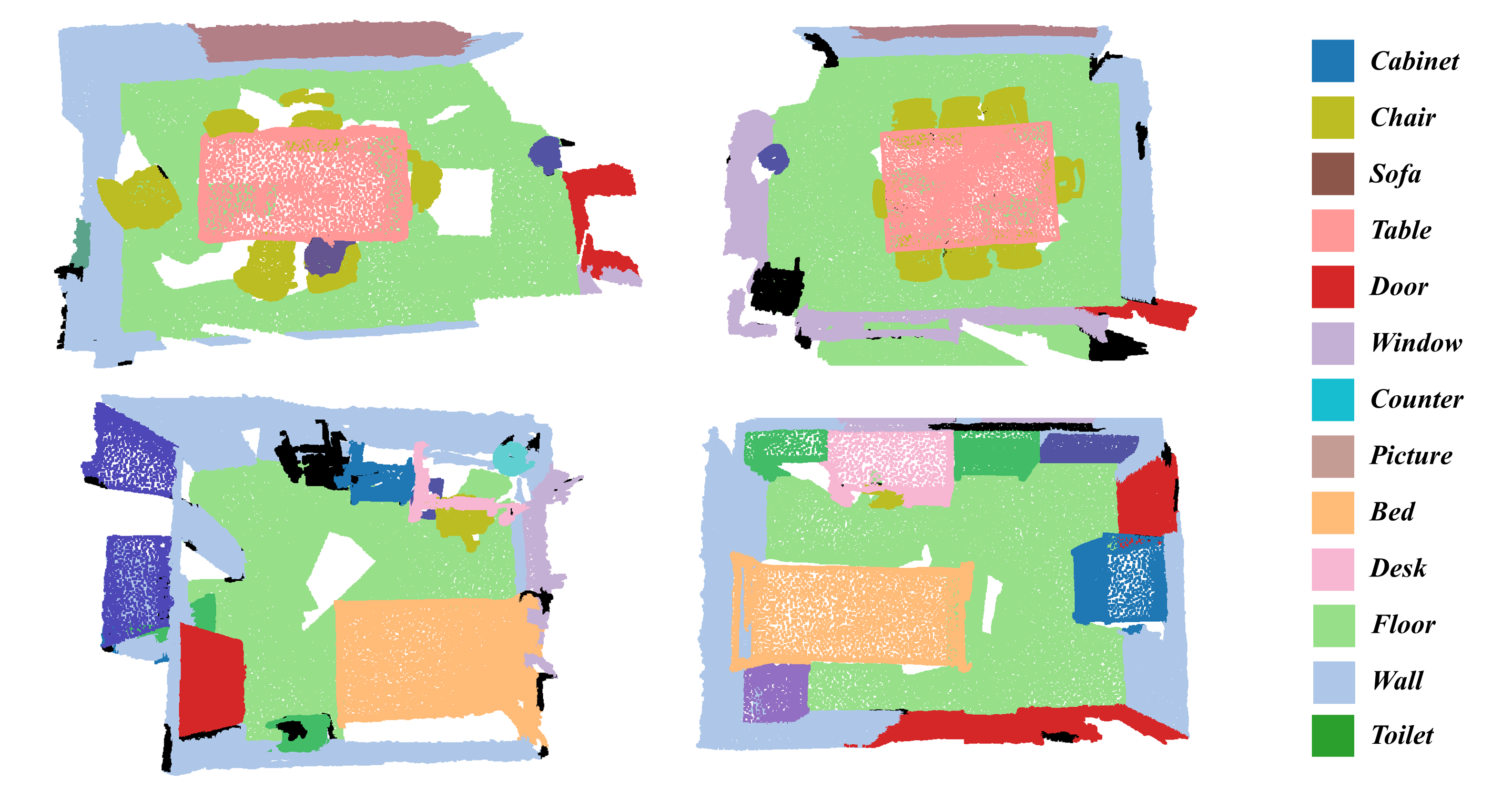}
\vspace{-5mm}
\caption{Qualitative results of semantic segmentation on ScanNet. }
\label{fig_viz_sem_seg}
\vspace{-4mm}
\end{figure}

To evaluate the semantic segmentation performance of our model, we use the ScanNet dataset for evaluation. Following prior works~\cite{dai2017scannet}, for each scene in ScanNet, a dense voxel grid with $0.02$m voxels was obtained by scene voxelization. Each voxel label was defined as the label with the most points in the corresponding voxel. We use the original training/validation/test data splits and report the mean Intersection over Union (mIoU) on the validation dataset (Val mIoU) and test dataset (Test mIoU). Our model is trained with 120 epochs, a batch size of 8, and augmented input data through random rotation and axis flip. We use the AdamW optimizer with a cosine learning rate scheduler.

We compare our model to some existing baselines. As shown in Table~\ref{tab_semseg}, our proposed model achieves the highest Val mIoU and Test mIoU compared to existing methods. This implies that moving 3D points toward the object center is an effective way to aggregate scene context and predict per point semantic labels. 
We also provide some visualizations of our results in Figure~\ref{fig_viz_sem_seg}. We can observe that the segmentation results generated by our model are compact and accurate.

\section{Conclusion}
\noindent We proposed a novel point-based 
voting step diffusion 
method for 3D object detection. Inspired by the advantages of generative models in modeling data distributions, we tackle the 3D object detection task with a noise conditioned score network by modeling the distribution of object centers and estimating the score function of the distribution to move from random points toward the vicinity of object centers.
To achieve this, we first generate a set of object center proposals and add normal noise to the generated object center proposals. We then estimate the score function of the object center distribution by predicting the added random noise using our proposed multi-scale score estimation module. Finally, we move the perturbed object center proposals toward the high-density region of the distribution according to the estimated score function. The denoised results are then aggregated to generate the final object proposals.
Extensive experiments on two large-scale indoor 3D scene datasets, ScanNet V2 and SUN RGB-D, demonstrate the effectiveness of our method and show that our method outperforms existing state-of-the-art point-based 3D object detection methods.

The performance of our proposed model is limited to some extent since it is trained on the point cloud dataset directly without any priors. Investigating diffusion models pre-trained on large scale common point cloud data and transferring them to 3D object detection framework is an exciting future direction.

\section{Acknowledge}
Professor Ajmal Mian is the recipient of an Australian Research Council Future Fellowship Award (project number FT210100268) funded by the Australian Government.

\bibliography{reference}{}
\bibliographystyle{IEEEtran}

\newpage

\vfill

\end{document}